\documentclass[eat,twocolumn]{jmlr}

\newcommand{\equal}[1]{{\hypersetup{linkcolor=black}\thanks{#1}}}
\usepackage{longtable}

\usepackage{booktabs}
\usepackage{xspace}
\usepackage{wrapfig}
\usepackage[font=small]{caption}
\usepackage[load-configurations=version-1]{siunitx} 

\theorembodyfont{\upshape}
\theoremheaderfont{\scshape}
\theorempostheader{:}
\theoremsep{\newline}

\jmlrvolume{}
\firstpageno{1}

\jmlryear{2022}
\jmlrworkshop{Machine Learning for Health (ML4H) 2022}

\title[Multi-modal Masked Autoencoders]{Multi-modal Masked Autoencoders Learn Compositional Histopathological Representations}

\author{%
\Name{Wisdom Oluchi Ikezogwo}$^{\equal{Equal contribution}}$ \Email{wisdomik@cs.washington.edu}\\
\Name{Mehmet Saygin Seyfioglu}$^{\footnotemark[1]}$ \Email{msaygin@cs.washington.edu}\\
\Name{Linda Shapiro}\Email{shapiro@cs.washington.edu}
\\ \addr University of Washington, Seattle, WA 98195, USA \\
}

\begin{document}

\maketitle

\begin{abstract}

Self-supervised learning (SSL) enables learning useful inductive biases through utilizing pretext tasks that require no labels. The unlabeled nature of SSL makes it especially important for whole slide histopathological images (WSIs), where patch-level human annotation is difficult. Masked Autoencoders (MAE) is a recent SSL method suitable for digital pathology as it does not require negative sampling and requires little to no data augmentations (DA). However, the domain shift between natural images and digital pathology images requires further research in designing MAE for patch-level WSIs. In this paper, we investigate several design choices for MAE in histopathology. Furthermore, we introduce a multi-modal MAE (MMAE) that leverages the specific compositionality of Hematoxylin \& Eosin (H\&E) stained WSIs. We performed our experiments on the public patch-level dataset NCT-CRC-HE-100K. The results show that the MMAE architecture outperforms supervised baselines and other state-of-the-art SSL techniques for an eight-class tissue phenotyping task, utilizing only 100 labeled samples for fine-tuning. Our code is available at \url{https://github.com/wisdomikezogwo/MMAE_Pathology}  

\end{abstract}

\section{Introduction}
\label{sec:intro}

In cancer pathology, tissue phenotyping is essential for learning accurate characterizations of histopathologic biomarkers within the tumor-immune microenvironments. The task of tissue phenotyping with computer vision is a complex endeavor, because the Giga-pixel resolution of whole slide images (WSIs) makes obtaining pixel-level and patch-level annotations challenging. Furthermore, the complexity in morphological phenotypes causes inter and intra-observer variability in tissue labeling.


Self-supervised learning (SSL) is an important technique for pre-training models without the need for large-scale labeled datasets. Instead, the model learns from the unlabeled data by means of a pretext task that utilizes a label generated from the data itself. The SSL framework presents a promising avenue for histopathology since human annotations are extremely costly. Recent SSL research on histopathology focused mostly on contrastive techniques \citep{yang2021self}, \citep{wang2021transpath}. However, one major drawback of contrastive SSL is that it depends on negative sampling. Negative sampling assumes a relatively good class balance within the dataset such that a sampled mini-batch would consist of images from different classes, which is often not the case for histopathology data. More recently, non-contrastive techniques like DINO \citep{caron2021emerging} are adapted to histopathology \citep{chen2022self}, that require no negative sampling. However, DINO still relies on DA techniques that are specifically designed for ImageNet (coarse-grained object-centric images), and thus is questionable in the context of histopathological images. Also, unlike natural images, WSI patches often have high granularity, as they consist of cells and other cellular structures scattered across a patch. Therefore, SSL on WSI patches is a fine-grained embedding learning problem, where designing DAs can be difficult \citep{xiao2020should}.

Recently, a new generative SSL framework Masked Autoencoders (MAE) was introduced \citep{he2022masked}. MAE forms autoencoder network using Vision Transformer (ViT) \citep{dosovitskiy2020image}, which masks random patches of a given input image and reconstructs the masked patches using the visible ones. MAE needs no negative sampling and does not depend on extensive DAs, thus it is very suitable for histopathology data. However, similar to the case of contrastive learning \citep{stacke2021learning}, heuristics in MAE designed for natural images do not necessarily transfer to histopathology, so more research is needed for adapting MAE to histopathology applications. We experimented with three different design choices for MAE: masking ratio, decoder depth, and patch size. Furthermore, we propose a multi-modal MAE (MMAE) that implicitly leverages human domain knowledge by incorporating H-stain, and E-stain images in addition to RGB, which helps the model to further leverage cross-modal interactions for better feature learning.

Our contributions are two-fold. First, we perform experiments to find the optimal MAE hyperparameters for tissue phenotyping using patch-level WSIs. Second, we propose the MMAE architecture that exploits human domain knowledge through the use of H$\&$E stain inputs, which provides compositional histo-morphological information in regards to cell and stroma. We compare our results against DINO, CS-CO and from-scratch baselines on a patch-level phenotyping task using a public dataset CT-CRC-HE-100K. Our results show that MAE outperforms contrastive/non-contrastive energy-based SSL techniques. Furthermore, we show that MAE's performance can be improved through the use of domain knowledge in the MMAE setting, which relies on a masking strategy that condenses all patch positions and samples uniquely from each modality. This allows for lighter masking ratios, which capture the granular nature of patch-level WSIs across all modalities.

\vspace*{-.3cm}

\section{Related Work}
\label{sec:Related}

Many SSL approaches have been developed for natural images \citep{pathak2016context},\citep{zhang2017split}, \citep{noroozi2016unsupervised}, \citep{gidaris2018unsupervised}. However, contrastive methods such as SimCLR \citep{yang2021self}, mostly dominated the field. Recently, DINO \citep{caron2021emerging} a non-contrastive method, which uses student-teacher distillation to learn part-whole relationships between the augmented views, was introduced, which outperforms SimCLR.

In histopathology domain, \citep{yang2021self} proposed CS-CO a contrastive SSL model with a novel augmentation method called stain vector perturbation, \citep{gildenblat2019self} implemented a contrastive loss based on the spatial distance of patches. Although these methods are shown to improve performance in detecting morphological phenotypes, they relied on DA methods designed for natural images, which do not always transfer well to pathology. DAs implicitly injects human domain knowledge, however, it is hard to engineer augmentations in histopathology as the downstream tasks show great dissimilarity \citep{stacke2021learning}. Finding good hard negatives also remains challenging, especially for patch-level WSIs, which are much more homogeneous compared to natural images. Lastly, negative sampling assumes a good class balance within the dataset which is not always valid for patch-level WSIs where there is an imbalance between morphological phenotypes \citep{chuang2020debiased}. Recently, non-contrastive
techniques, such as DINO, are also adapted on histopathology data \citep{chen2022self}. However, DINO still relies on extensive DA techniques, and also requires a lot of hyperparameters to be tuned, which makes it cumbersome to train. Therefore, a framework that does not rely on extensive augmentation techniques and negative sampling would be a better fit for histopathology data, and MAE perfectly fits in this role.
\begin{figure*}[ht!]
\centering
\includegraphics[width=\textwidth,height=7cm]{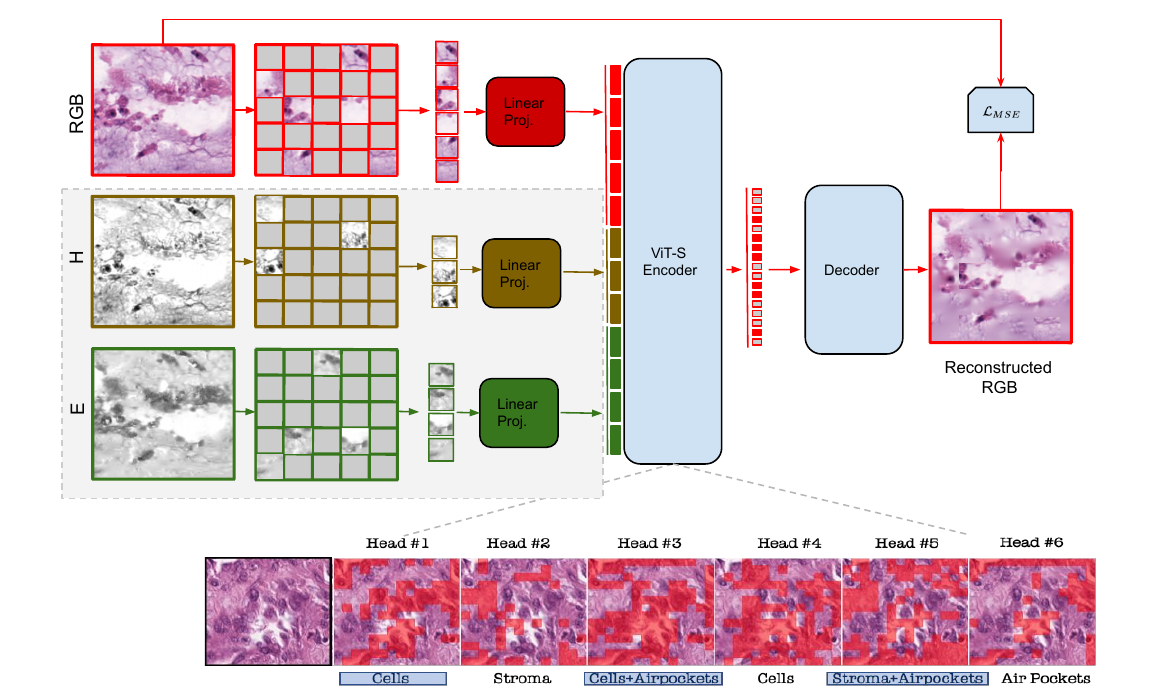}
  \caption{\textbf{MMAE architecture}  During pre-training, a small random subset of patches is masked out. \textbf{Interpretation} Thresholded visualization of multi-head attention weights shows that MMAE learns to localize different histo-morphological features like cells in \#1, 3 and 4, stroma tissue in \#2 and 3, and regions of fat/air-pockets in \#3, 5 and 6, provides empirical evidence that MMAE can capture important inductive biases about histopathology tissue compositionality.}
  \label{fig:MMAE}
\end{figure*}

\section{Methodology}




\subsection{Masked Autoencoders} \label{mae_section}
Masked Autoencoders (MAE) is a generative SSL technique that yields state of the art performance in ImageNet benchmarks \citep{he2022masked}. MAE randomly masks a portion of the patches of the input image, feeds the unmasked (visible) patches to its ViT encoder, and then reconstructs the masked patches using a lightweight ViT decoder by optimizing a mean squared error loss $\mathcal{L}_{MSE}$.

\textbf{Masking.}
An input image is patchified by non-overlapping patches. Then a fixed number of randomly selected patches are sampled without replacement and masked. 

\textbf{Encoder and Decoder} MAE's backbone encoder is a ViT, which takes non-masked input patches. Each visible patch is first flattened into a vector embedding by a linear projection, then a 2D sinusoidal position embedding is added to it. The resulting vectors are then processed by the Transformer blocks. The decoder also consists of Transformer blocks, with the task of reconstructing the masked patches using the latent variables created by the encoder. The decoder is removed in the downstream fine-tuning.


\subsection{Multi-Modal MAE}\label{mmae_section}

Inspired by \citep{bachmann2022multimae}, we propose a Multi-Modal MAE (MMAE) that leverages H\&E stains in addition to RGB inputs. The overall architecture of our MMAE differs from \citep{bachmann2022multimae} as we do not do multi-task reconstruction (no reconstruction for H\&E inputs), so we have a single decoder that is dedicated to RGB reconstruction. By incorporating H\&E stains, we are infusing the domain knowledge, since H\&E stains highlight different histomorphological features. Thus using the tokens from H\&E allows us to further reduce masking of RGB tokens, since it is nontrivial to reconstruct the RGB tokens using H\&E tokens. The overall architecture is shown in Fig \ref{fig:MMAE}. 




\textbf{MMAE Masking.} We use a Dirichlet distribution to determine the amount of visible tokens for each modality by sampling from $Dir(\alpha)$ distribution, which is controlled by the concentration parameter $\alpha > 0$, where equal $\alpha$'s between modalities mean a uniform sampling. A difference in our sampling strategy from that of  \citep{bachmann2022multimae} is that they sample randomly across their input modalities, which can result in the same patch position being taken in more than one modality. However, since WSIs are highly granular, we need visible tokens to be complementary, thus information coming from additional modalities must be ensured. To provide that, we implemented the \textit{mask-one} strategy, which ensures that unique tokens will be selected across all modalities. (See Appendix for details.) 

\textbf{MMAE Encoder and Decoder.} MMAE utilizes separate linear projections for each input, then adds position embeddings on top of each patch vector, and concatenates the resulting vectors. Concatenated patch vectors are then fed into the same Transformer architecture used as given in the MAE encoder. No modality specifying token is used explicitly. The decoder aims to reconstruct only the RGB input, so unlike \citep{bachmann2022multimae}, we use one decoder. We add a single cross-attention layer in the decoder to stimulate cross-modal interaction between the latent variables from each input modality.

\vspace{-0.3mm}
\section{Experiments}

We designed a set of experiments utilizing ViT-S as our encoder. For MAE, we experimented with three different hyperparameters: patch size, decoder depth, and masking ratio, keeping two fixed and varying the remaining one. For MMAE, we kept the best hyperparameters from the MAE experiments for decoder depth and patch size, and experimented with different masking and sampling strategies. We tried a) equal masking for all modalities, b) heavy masking on RGB and light masking on H\&E, and c) heavy masking on H\&E and light masking on RGB.

\textbf{Dataset.}
We trained and validated our models on the NCK-CRC-100K dataset of $100,000$ histological images of human colorectal cancer and healthy tissue, extracted as $224$$\times$$224$ patches at 0.5 microns per pixel (MPP) with 8 classes. The final sizes of the training and test sets are $89,434$ and $6333$, respectively, and the evaluation measure is the eight-class classification accuracy of the test set; see Appendix for details.
\textbf{H\&E Stain Separation.}
Different dyes are employed in histology to emphasize various types of tissue components, which might be considered domain-specific knowledge implicit in histopathological images \citep{alturkistani2016histological}. In our case, cell nuclei will be stained purple by hematoxylin (H stain image), while the extracellular matrix and cytoplasm will be stained pink by eosin (E stain image) in the commonly used H\&E stain. Similar to the work by \citep{yang2021self} we use the Vahadane approach \citep{vahadane2016structure}. See Appendix for details.


\textbf{Pre-training and Downstream Fine-tuning Details.} We pre-train the models for $1600$ epochs using four NVIDIA A4000 GPUs with automatic mixed precision enabled. See Appendix for the full list of hyperparameters. For the downstream, we fine-tune for the 8-class classification for $100$ epochs, using 5-fold cross validation(CV). Also, instead of using the entire training set, we randomly select 100 and 1000 samples from the training set, and fine-tune with this group to better represent each method's representation capacity. Note that for the MMAE fine-tuning, we only utilize the RGB inputs. We compare our results against DINO, and CS-CO. For DINO, we use the settings given in \citep{chen2022self}, and for CS-CO, we report results directly from \citep{yang2021self}.

\vspace{-2mm}
\section{Results}

\textbf{MAE Results}. In all our experiments, we fine-tuned the pre-trained MAE model using N = 100, and N = 1000 samples in two different experiments shown in Fig \ref{fig:all_hyperparam}. For \textbf{Decoder Depth}, we kept the masking ratio at 75$\%$ and patch size at 16$\times$16 and experimented for decoder size of 2, 6, 10.. For the \textbf{Patch Size}, we experimented with the patch sizes of 8$\times$8, 14$\times$14, and 16$\times$16 by keeping the decoder depth at 2, and masking ratio at 75$\%$. Lastly, for the \textbf{Masking Ratio}, we kept the patch size at 16$\times$16 and decoder depth at 2 and experimented with different masking ratios.



\begin{figure}[h]
\floatconts
  {fig:all_hyperparam}
  {\caption{Results for Hyperparameter Variations}}
  {\includegraphics[width=1\linewidth, height = 4.4cm]{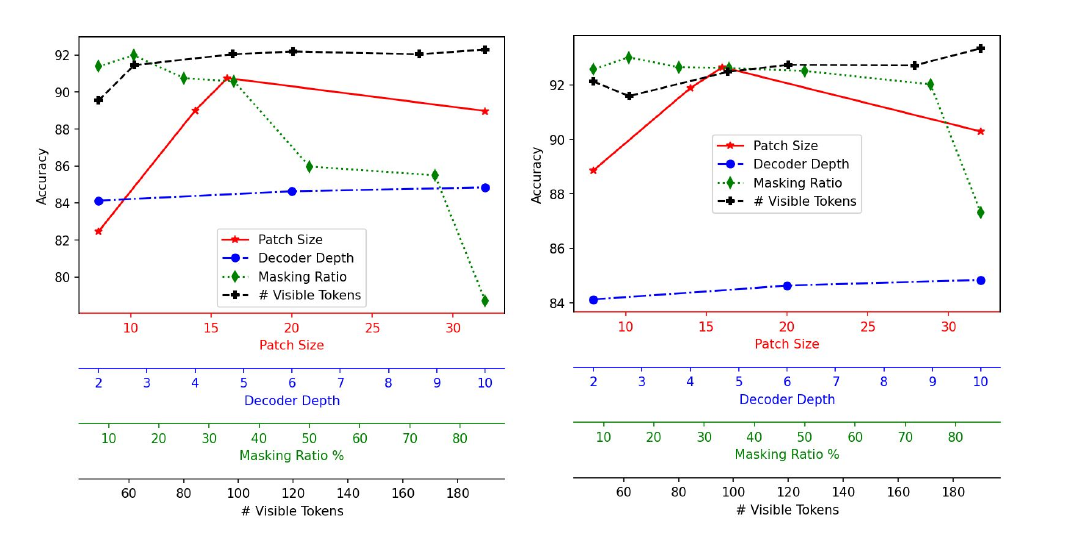}}
  \vspace*{-2mm}
\end{figure}
\vspace{-1mm}






\textbf{MMAE Results}.First, we experimented with different masking strategies on different modalities based on $Dir(\alpha)$. Results show that heavy masking on E\&H, and light masking on RGB works best ($\alpha_{\text{RGB}} = 0.8$, $\alpha_{\text{H}} = \alpha_{\text{E}} = 0.1$) Then we experimented with different masking ratios using the optimal masking strategy. We found that the MMAE allows us to reduce the masking ratio further.

\textbf{Results Compared Against Other SSL Methods.} Results in Table \ref{tab:sota} indicate that both MAE and MMAE outperform other SSL methods and from-scratch baselines for the eight-class tissue phenotyping task. We also see that MMAE slightly outperforms MAE, as it allows us to reduce the overall masking ratio to better capture the granular nature of patch-level WSIs, while increasing the individual masking applied to RGB, thereby alleviating the risk of trivial interpolation.

\begin{table}[hbtp]
\footnotesize
\floatconts
  {tab:sota}
  {\caption{Accuracy and PRAUC results (shown as Accuracy/PRAUC) obtained with 5-fold CV}}
  {\begin{tabular}{l|lll}
  \toprule
  \bfseries Models & \bfseries n = 100 & \bfseries n = 1000 & \bfseries n = all\\
  \midrule
  DINO & 89.6/91.1 & 92.6/95.2 & 84.9/93.3\\
  MAE & 92/94.6 & 93/95.6 & 88.1/94.5\\
  MMAE & 92.3/94.1 & 93.3/96 & 88.4/94.8\\
  \bottomrule
  \end{tabular}}
  \vspace*{-.3cm}
\end{table}

\textbf{Pre-training Performance Comparison Against DINO.} We set up additional experiments to evaluate the pre-training performance of MMAE by nearest neighbor matching against DINO, the runner-up of our benchmarks. MMAE outperforms DINO in the nearest neighbor matching setting, where DINO yields around 87\% for 10 nearest neighbors (NN) and 87.45\% for 20 NNs, whereas MMAE yields 90.19\% and 90.94\%, for NN=10 and NN=20, respectively.



\vspace*{-.2cm}
\section{Discussion}

\textbf{MAE}. For \textbf{Masking Ratio}, our results indicate that heavy masking performs poorly for patch-level WSIs as opposed to natural images. For coarse-grained natural images, it is easy for the model to extrapolate using fewer patches, since a single object can be inferred utilizing a limited signal. However, patch-level WSIs have a higher degree of granularity. Thus, the model requires more visible patches for learning the cellular structures. Also, masking too lightly yields poor results, as having too many visible adjacent patches leads to trivial interpolation, which aligns with the findings in natural images. Interestingly, optimal masking ratios for patch-level WSIs are similar to masked language modeling in natural language processing, where the typical masking ratio is also around 15\%. For \textbf{Decoder Depth}, we found that increasing the depth brings mixed results. Though it provides a slight performance boost when fine-tuned with N=100 samples, it degrades performance with N=1000 samples. We argue that, given the capacity, the decoder may be leaning towards learning shortcut features that impairs feature learning on the encoder. So the optimal decoder depth should be shallow for patch-level WSIs.  Lastly for the \textbf{Patch Size}, we found that the optimal patch size to be 16$\times$16, the same as for natural images. We argue that if cell-sized or sub cell-sized patches are employed, as in the case of the 8$\times$8 patches, the reconstruction task becomes a trivial interpolation. 





\textbf{Multi-modal MAE.} Results indicate that light masking on RGB and heavy masking on E\&H outperforms other masking strategies. This outcome was expected because the pre-training task is to reconstruct the RGB input and fine-tuning objective utilizes only the RGB input. Also, the results show that MMAE outperforms other SSL methods. We observe that MMAE relaxes the masking trade-off, as it allows us to use more visible tokens overall than MAE. However, note that MMAE uses less visible RGB tokens, so the extra tokens come from E\&H. This helps the model better capture the granular nature of the WSIs without risking trivial interpolation. Since E\&H images encode different information and are disparate from the RGB image, reconstructing RGB tokens using E\&H is not straightforward. By using fewer RGB tokens than MAE (without the expense of signal loss), MMAE better captures the cell structure, and thus outperforms in both fine-tuning and nearest neighbor matching tasks.
\vspace{-.5mm}
\textbf{Limitations and Conclusions.} There are some limitations of this work. First, we did not perform a comprehensive hyperparameter search to determine the combinations of design choices for MAE, as pre-training requires too much computation. Second, NCK-CRC-100K is a relatively small dataset, so the experiments must be re-conducted on a larger patch-level dataset to validate the effectiveness of MMAE. Our work shows the efficacy of generative pre-training with MAE compared to other SSL techniques. We show that MAE benefits from low masking ratios for patch-level WSIs. Lastly, we show that the performance of MAE can be improved by incorporating H\&E stains in the MMAE setting, which provide compositional signals that hold cellular information and allow us to further reduce the masking ratio to better capture the granular nature of patch-level WSIs.

\bibliography{pmlr-sample}

\clearpage

\appendix

\section{ }\label{apd:first}
\textbf{Dataset Details} \label{data}
NCK-CRC-100K is annotated with the following non-overlapping tissue classes: adipose (Adi), background (Back), debris (Deb), lymphocytes (Lym), mucus (Muc), smooth muscle (Mus), normal colon mucosa (Norm), cancer-associated stroma (Str), colorectal adenocarcinoma epithelium (Tum). We experiment on CRC-100K without Macenko SN. When evaluating visual representation learning methods, we don't use images from the background (BACK) class for training and testing. This is because they are always non-informative, and can be easily predicted using simple threshold based approaches \citep{yang2021self, chen2022self}.

\textbf{Stain separation formulation}
Elaborating mathematically, let $I \in \mathbb{R}^{m \times n}$ be the RGB intensity matrix, $V \in \mathbb{R}^{m \times n}$ be the relative optical density, $W \in \mathbb{R}^{m \times r}$ be the stain color matrix, and $H \in \mathbb{R}^{r \times n}$ be the stain concentration matrix for an H\&E stained image, where $m = 3$ for RGB images, $r$ is the number of stains, and $n$ is the number of pixels. The relationship between $V$ and $H$,$W$ can be stated as $V = \log \frac{I_0}{I} = WH$, where $I_0 = 255$ for 8-bit RGB pictures, according to the Beer-Lambert law. Then, $W$ and $H$ can be estimated by solving the sparse non-negative matrix factorization problem as Eq. (1) proposed by \citep{vahadane2016structure}.

\begin{equation}
\begin{split}
    \min_{W,H} \frac{1}{2} \| V - WH \|^2_F + \lambda \sum^r_{j=1} \| H(j, \colon)\|_1 \\
    \ni W, H \geq 0,  \quad \|W(\colon, j)\|^2_2 = 1
\end{split}
\end{equation}

The H channel and E channel images $I_h$ and $I_e$ can be recovered as $I_h = I_0 \exp (-H[0, \colon])$ and $\quad I_e = I_0 \exp (-H[1, \colon])$, respectively, using the estimated stain concentration matrix $H$. See Fig \ref{fig:MMAE} for RGB, H and E images.

\textbf{Sampling Strategy}
Depicted in Fig \ref{fig:sampling}-a, \textit{Mask-all} effectively treats each patch position in all three modalities equally, thus making the sampling space equal to the total number of patches available ($3 * M$). Using our best MAE mask ratio of 15\% would mean encoding $\approx$500 patches, which would lead to inefficiencies due to the quadratic memory and compute constraints of self-attention. However, with \textit{Mask-one}, due to the compositional nature of H and E images, we can reduce our sampling space to $M$ by sampling each unique patch position just once \ref{fig:sampling}.


\begin{figure}[h!]
\floatconts
  {fig:sampling}
  {\label{fig:sampling}}
  {\includegraphics[width=1\linewidth]{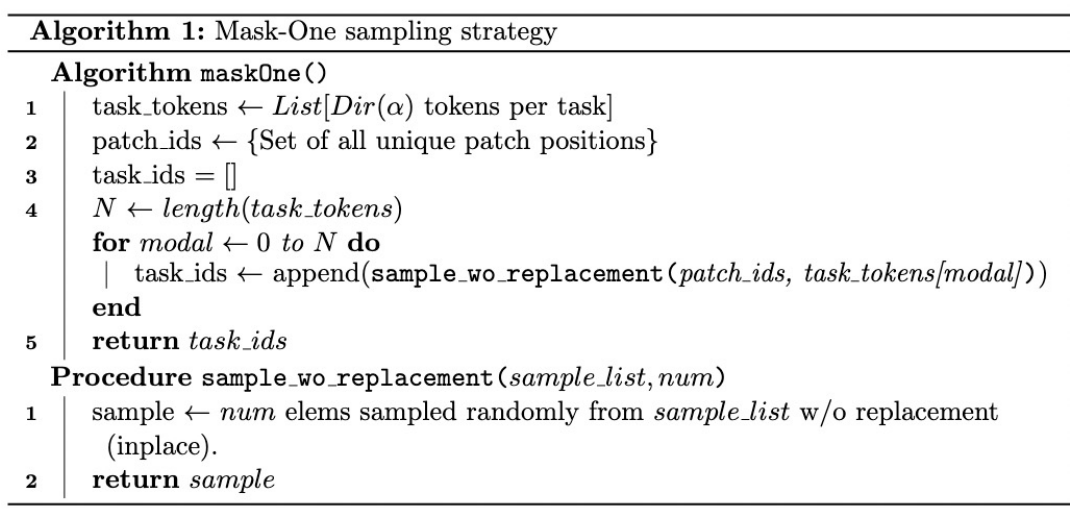}}
\end{figure}

\textbf{Training details}
In Table \ref{table:train_dets} we report the changes to the default pre-training setting in \citep{bachmann2022multimae}.
\begin{table}[!htp]
    \centering
    \scriptsize	
    \setlength\tabcolsep{1.5 pt} 
    \begin{tabular}{l|l} 
        \toprule
        Hyperparameters & Value \\
        \midrule
        \textbf{Pre-training} & \\
        \midrule
        Base lr & 1e-4 \\
        Batch size & 312 \\
        Augmentation & RandomResizedCrop (scale: 0.8, 1.0) \\
        Num. global tokens & 1 \\
        Epochs & 1600 \\
        Sampling $\alpha$ & $\alpha_{RGB}$: 8, $\alpha_{H}$ :1, $\alpha_{E}$ :1 \\ 
        Norm pix loss & True \\
        Patch size & 8, 14, 16 \\
        Decoder depth & 2, 6, 10 \\
        \midrule
        \textbf{Fine-tuning} & \\
        \midrule
        Base lr & 3e-3 \\
        Weight decay & 6e-5 \\
        Batch size & 96 \\
        Epochs & 100 \\
        \bottomrule
    \end{tabular}
    \caption{\textbf{Pre-training settings.}}
    \label{table:train_dets}
\end{table}

\vspace*{-.3cm}
\begin{figure*}[b!]
\floatconts
    {fig:sampling}
    {\caption{(a) MMAE sampling strategies: In this simple (4x4) grid setting, the grids show selected patch positions for each modality (b) We provide more attention visualizations for all 8-classes.}}
    {\label{fig:sampling}}
    {%
    \subfigure{%
      \label{fig:beav}%
      \includegraphics[width=1\linewidth, height = 5cm]{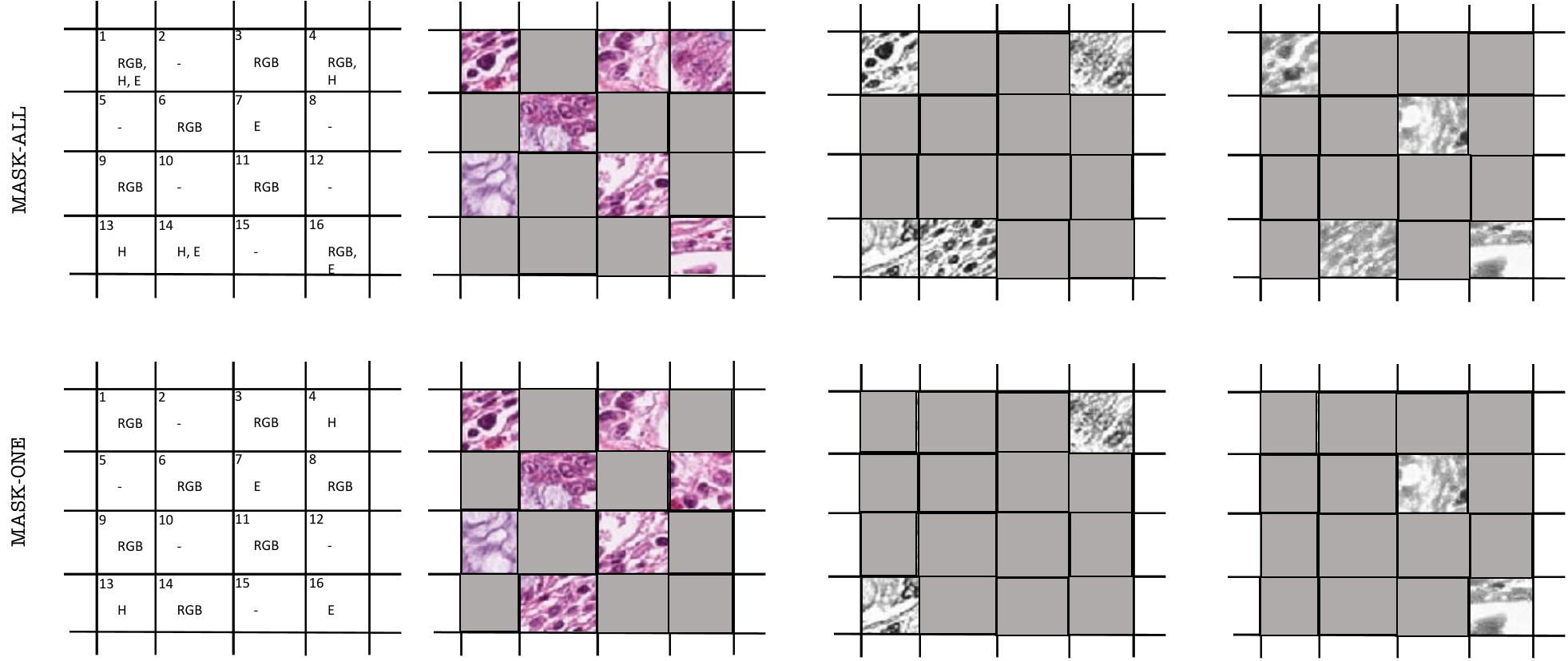}}%
    \qquad %
    \subfigure{%
      \label{fig:nck}%
      \includegraphics[width=1\linewidth, height = 4.5cm]{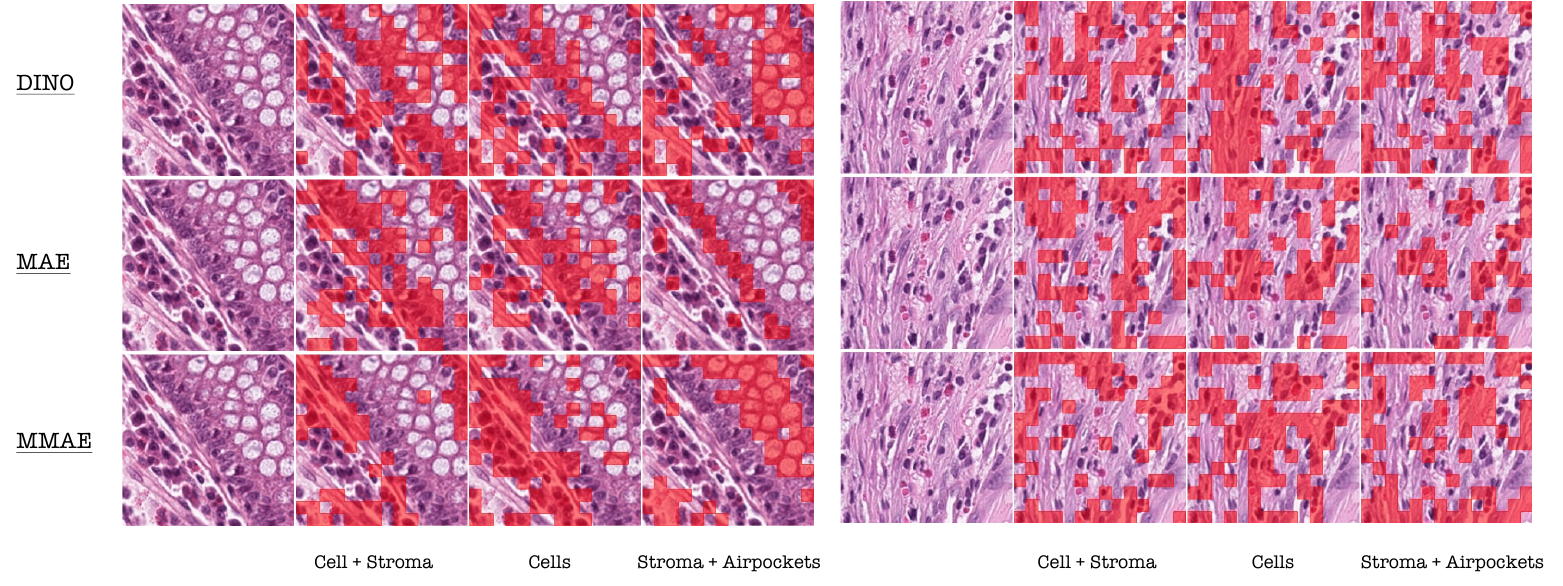}}
  }
\end{figure*}

\begin{figure*}[b!]
\floatconts
  {fig:umaps}
  {\caption{Using the default UMAP parameters ($neighbors = 15, dist =0.1$) we obtained the 2D scatter plots for DINO, MAE, and MMAE using the [CLS] tokens of each model. For all 8-classes. Note that MMAE clusters them into separable groups in comparison to others}}
  {\includegraphics[width=1\linewidth, height = 5cm]{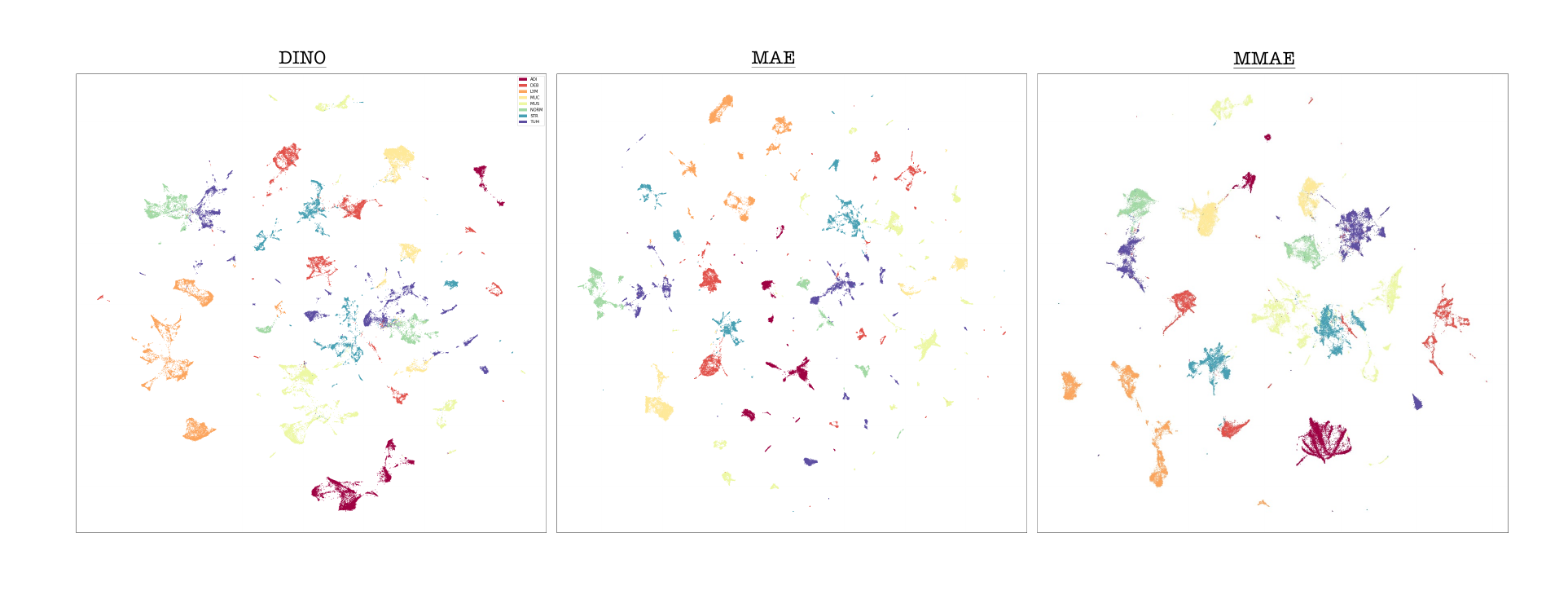}}
\end{figure*}

\clearpage . \let\clearpage\relax



\end{document}